\documentclass{article}

\PassOptionsToPackage{numbers, compress}{natbib}

\usepackage[preprint]{nips_2018}

\usepackage[utf8]{inputenc} 
\usepackage[T1]{fontenc}    
\usepackage{hyperref}       
\usepackage{url}            
\usepackage{booktabs}       
\usepackage{amsfonts}       
\usepackage{nicefrac}       
\usepackage{microtype}      

\usepackage{multirow}
\usepackage[inline]{enumitem}
\usepackage{graphicx}
\usepackage{bm}
\usepackage{amsmath}
\usepackage{xcolor}

\def\x{\bm{x}}
\def\X{\bm{X}}
\def\z{\bm{z}}

\def\bmu{\bm{\mu}}

\def\L{{\cal L}}
\def\N{{\cal N}}
\def\E{{\mathbb E}}

\title{Disentangling by Partitioning: A Representation Learning Framework for Multimodal Sensory Data}

\author{
  Wei-Ning Hsu, and James Glass \\
  Computer Science and Artificial Intelligence Laboratory\\
  Massachusetts Institute of Technology\\
  Cambridge, MA 02139, USA \\
  \texttt{\{wnhsu,glass\}@mit.edu} \\
}

\begin{document}

\maketitle

\begin{abstract}
Multimodal sensory data resembles the form of information perceived by humans for learning, and are easy to obtain in large quantities.
Compared to unimodal data, synchronization of concepts between modalities in such data provides supervision for disentangling the underlying explanatory factors of each modality.
Previous work leveraging multimodal data has mainly focused on retaining only the modality-invariant factors while discarding the rest. 
In this paper, we present a partitioned variational autoencoder (PVAE) and several training objectives to learn disentangled representations, which encode not only the shared factors, but also modality-dependent ones, into separate latent variables.
Specifically, PVAE integrates a variational inference framework and a multimodal generative model that partitions the explanatory factors and conditions only on the relevant subset of them for generation.
We evaluate our model on two parallel speech/image datasets, and demonstrate its ability to learn disentangled representations by qualitatively exploring within-modality and cross-modality conditional generation with semantics and styles specified by examples.
For quantitative analysis, we evaluate the classification accuracy of automatically discovered semantic units.
Our PVAE can achieve over 99\% accuracy on both modalities.

\end{abstract}

\section{Introduction}
To build an artificial intelligence that learns and thinks like humans, as suggested in~\cite{bengio2013representation, lake2017building}, one needs to design a machine that can understand the world. 
Such understanding can only be achieved by learning to identify and disentangle the underlying explanatory factors from the observed low-level sensory data, for example, word and speaker identity from speech.
This process is also referred to as representation learning, and is one of the fundamental problems in machine learning.
Aside from cognitive scientists' interest, interpretable and disentangled representations have also been proven useful in a wide variety of tasks, such as zero-shot learning, novelty detection, and transfer learning~\cite{lake2017building,edwards2016towards,hsu2018extracting}, where humans excel and supervised models fail. 

Variational autoencoders (VAEs)~\cite{kingma2013auto, rezende2014stochastic} provide a general and powerful framework for learning representations by combining neural networks and probabilistic generative models~\cite{koller2009probabilistic}:
the causalities between variables are specified by the probabilistic graphical model, 
and the complex non-linear conditional relationships are captured by the neural networks.
Learning in this scenario corresponds to fitting the model parameters such that the likelihood of the observed dataset is maximized, and representations refer to the inferred values of the latent variables of the data. 
This framework has demonstrated great success in learning representations directly from raw sensory data, including images~\cite{kingma2013auto}, speech~\cite{hsu2017learning}, and videos~\cite{denton2018stochastic}.

In this paper, we investigate the task of discovering explanatory factors from multimodal sensory data, such as parallel images and speech recordings, resembling what humans perceive during learning.
Compared to unimodal data, synchronization between different modalities with such data provides supervision for reasoning about the underlying generative process and disentangling shared explanatory factors from the rest.
Previous work that utilizes multimodal sensory data has been mainly focused on the scenario where the objective is to extract the shared explanatory factors while discarding the rest~\cite{harwath2015deep, harwath2016unsupervised, kamper2017visually, leidal2017learning, senocak2018learning}.
Such representations are only useful for a subset of downstream tasks, but cannot be applied to tasks that demand those discarded factors.
Our goal on the other hand is to learn representations for not only the shared explanatory factors, but also the modality-dependent factors, and to encode them in different latent variables for disentanglement and interpretability.
We present a partitioned generative model for multimodal data where each modality involves one modality-invariant latent semantic variable and one modality-dependent latent style variable.
By integrating this model with unimodal and multimodal inference models, we propose a novel partitioned variational autoencoder (PVAE) and several training objectives for learning variable-level disentangled representations and promoting disentanglement.
Our model is evaluated on two multimodal spoken/written digit datasets.
Both quantitative and qualitative results verify the usefulness of our generative model and demonstrate the ability to separate semantic and style information.
In particular, PVAE is also capable of automatically discovering the number of digit classes 
and achieves over $99\%$ classification accuracy on both modalities.

\section{Disentangling by partitioning}
In this section, we introduce a partitioned latent variable model for a generative process of multimodal data.
By integrating such a partitioned graphical model with the neural variational inference framework, we then propose a novel partitioned variational autoencoder (PVAE) and training objectives.

\subsection{Proposed framework: partitioned variational autoencoder (PVAE)}
Consider a multimodal dataset, where each sample is a tuple of paired data in different modalities (e.g., a collection of images and their spoken captions in different languages~\cite{harwath2018vision}, or synchronized audio and image streams from YouTube~\cite{arandjelovic2017objects}).
We argue that generation of such data involves two types of factors: one captures the concept that is jointly described by all the modalities, while the other specifies the residual factors for a particular modality.
Therefore, instead of conditioning all modalities on a single shared latent variable, 
we propose to \textit{partition the set of explanatory factors, and condition the generation of each modality only on the relevant subset.}

\subsubsection{Partitioned generative network for multimodal data}
\begin{figure}[ht]
  \centering
  \includegraphics[width=.8\linewidth]{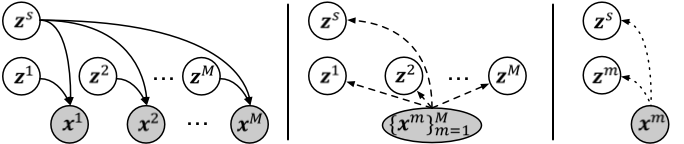}
  \caption{Graphical illustration of the proposed graphical models, where gray nodes denote the observed variables, and white nodes denote the latent variables. (left) the generative network, $p(\cdot | \cdot)$. (middle) the multimodal inference network, $q(\cdot | \cdot)$. (right) the unimodal inference network, $r(\cdot | \cdot)$.}
  \label{fig:pgm}
\end{figure}

Given a dataset $\mathcal{D} = \{ \{ \x_i^m \}_{m=1}^M \}_{i=1}^N$ of $N$ i.i.d. $M$-modality samples, where $m$ indexes modalities and $i$ indexes samples. 
Note that the i.i.d. assumption holds among samples, but not across modalities.
We assume that each sample $(\x^1, \x^2, \cdots, \x^M)$ is generated from some random process involving one \textit{latent semantic variable} $\z^s$ and $M$ \textit{latent style variables} $(\z^1, \z^2, \cdots, \z^M)$.\footnote{We drop the subscript $i$ whenever it is clear that we are referring to terms associated with a single sample.} 
Consider the following generative process depicted in Figure~\ref{fig:pgm} (left):
\begin{enumerate*}[label=(\arabic*)]
  \item a latent semantic variable $\z^s$ is drawn from a semantic prior distribution $p(\z^s)$;
  \item a latent style variable $\z^m$ is drawn from its associated modality-specific prior distribution $p(\z^m)$ for each modality $m \in \{ 1, \cdots, M \}$; 
  \item an observed variable in each modality is drawn from a modality-specific conditional distribution $p(\x^m | \z^m, \z^s)$.
\end{enumerate*}
The joint probability for a multimodal sample is formulated as:
\begin{equation}
  p( \{ \x^m \}_{m=1}^M, \{ \z^m \}_{m=1}^M, \z^s) = p(\z^s) \prod_{m=1}^M p(\z^m) p( \x^m | \z^m, \z^s).
\end{equation}
Specifically, we assume all the prior distributions to be centered isotropic Gaussian $\N(\bm{0}, \bm{I})$ with no trainable parameters.
The conditional distribution of each modality is assumed to be a diagonal Gaussian, whose mean and variance are parameterized by neural networks that take the corresponding latent variables as input.
The vector $\theta$ denotes the trainable parameters in this generative model.

\subsubsection{Multimodal and unimodal inference networks}
For the multimodal variational inference, we consider the model in Figure~\ref{fig:pgm} (middle):
\begin{equation}
  q( \{ \z^m \}_{m=1}^M, \z^s | \{ \x^m \}_{m=1}^M ) = q(\z^s | \{ \x^m \}_{m=1}^M ) \prod_{m=1}^M q(\z^m | \{ \x^m \}_{m=1}^M ).
\end{equation}
The posterior over each latent variable is a diagonal Gaussian distribution, whose mean and variance are parameterized by neural networks that take a multimodal sample as input. The vector $\phi$ is used to denote the collection of trainable parameters in this inference model.

We now derive inference models for unimodal data, which is desirable for the case when a testing sample contains only one modality.
Given a unimodal sample $\x^l$ from the $l$-th modality, the exact posterior over all latent variables can be derived as:
\begin{align}
  p( \{ \z^m \}_{m=1}^M, \z^s | \x^l ) &= \dfrac{ p( \x^l | \{ \z^m \}_{m=1}^M, \z^s) p(\z^s) \prod_{m=1}^M p(\z^m) }{ p(\x^l) } \\ 
  &= \dfrac{ p( \x^l | \z^l, \z^s) p(\z^s) \prod_{m=1}^M p(\z^m) }{ p(\x^l) } \label{eq:pos_ci} \\ 
  &= p(\z^l, \z^s | \x^l ) \prod_{ \substack{m \in \{1, \cdots ,M\} \\ m \neq l} } p(\z^m), \label{eq:pos_final}
\end{align}
where Eq.~\eqref{eq:pos_ci} results from the conditional independence assumption based on the generative network.
Equation~\eqref{eq:pos_final} implies that only the latent semantic variable $\z^s$ and the corresponding latent style variable $\z^l$ can be inferred from a unimodal sample $\x^l$. 
Hence, the following unimodal variational inference network is considered:
\begin{equation}
  r(\z^l, \z^s | \x^l) = r(\z^l | \x^l) r(\z^s | \x^l),
\end{equation}
where each term on the right hand side is again a diagonal Gaussian with mean and variance parameterized by neural networks, whose parameters are collectively denoted with the vector $\psi$.

\subsection{Training objectives}
The combination of the partitioned generative networks with multimodal and unimodal inference networks is termed the \textit{Partitioned Variational Autoencoder} (PVAE), where the inference networks and the generative network can be regarded as probabilistic \textit{encoders} and \textit{decoder}, respectively.
In this section, we discuss several objective functions for training PVAE models to promote disentanglement.

\subsubsection{Variational lower bound}
Let $\X = \{ \x_i^m \}_{m=1}^M$ denote a multimodal sample, for the multimodal inference network. A variational lower bound on the marginal log likelihood,
\begin{align}
  \L (\theta, \phi; \X ) = \sum_{m=1}^M \big[ \E_{q(\z^m, \z^s | \X )} [ \log p(\x^m | \z^m, \z^s) ] - D_{KL}( q(\z^m | \X) || p(\z^m) ) \big] - D_{KL}( q(\z^s | \X) || p(\z^s) ), \nonumber
\end{align}
consists of $M$ negative expected reconstruction errors, and $M+1$ negative KL-divergence terms.
The reconstruction terms and the KL-divergence terms act as two competing forces to the multimodal encoder: 
while the former benefits from having its conditioned latent variables encode as much information as possible about the corresponding observed variable, 
the KL-divergence prefers an uninformative posterior that is the same as the prior regardless of observed variables.
By encoding the shared explanatory factors in the latent semantic variable $\z^s$, only one KL-divergence term $D_{KL}( q(\z^s | \X) || p(\z^s) )$ is penalized, instead of $M$ terms $\sum_{m=1}^M D_{KL}( q(\z^m | \X) || p(\z^m) )$, while providing the same information to the decoders.

\subsubsection{Coherence between multimodal and unimodal inference}
The above variational lower bound only optimizes the partitioned generative networks and the multimodal inference network.
Up until now, we assume that we have access to all the modalities. In the case when we only have a single modality, we would like to disentangle the factors similarly to the multimodal case. 
To do this, we propose an objective called \textit{multimodal-unimodal coherence}:
\begin{equation}
  CH(\phi, \psi; \X) = \sum_{m=1}^M -D_{KL} ( q(\z^m, \z^s | \X) || r(\z^m, \z^s | \x^m) ),
\end{equation}
which aims to minimize the KL-divergence between the multimodal posterior and the unimodal posterior over the latent semantic variable and the corresponding latent style variable.

\subsubsection{Cross-modality semantic contrastiveness}
While the variational lower bound $\L (\theta, \phi; \X )$ encourages shared explanatory factors to be represented with the latent semantic variable, it does not discourage non-shared explanatory factors from being encoded in the latent semantic variable.
Suppose we have two multimodal samples, $\X = \{ \x^m \}_{m=1}^M$ and $\tilde{\X} = \{ \tilde{\x}^m \}_{m=1}^M$, whose semantic differ from each other. 
Ideally, we would like the posteriors over the latent semantic variable inferred by different modalities from the same sample to be similar, and to be dissimilar when inferred from different samples.
Based on this intuition, we propose a hinge loss-based objective named \textit{cross-modality semantic contrastiveness} as follows:
\begin{equation}
  CM(\phi, \psi; \X, \tilde{\X}) = -\dfrac{1}{M} \sum_{\substack{m,m' \in \{ 1, \cdots, M \} \\ m \neq m'}} 
  \max(0, t - 
  	k( \bmu_{\z^s | \x^{m'}}, \bmu_{\z^s | \x^{m}} ) + 
    k( \bmu_{\z^s | \x^{m'}}, \bmu_{\z^s | \tilde{\x}^{m}} ) ),
\end{equation}
where $\bmu_{\z^s | \x^m}$ denotes the unimodal posterior mean of $\z^s$ inferred from $\x^m$ (mean of $r(\z^s | \x^m)$), $k(\cdot, \cdot)$ is a kernel function measuring the similarity between two posterior means, and $t$ denotes the margin.
Specifically, we use a radius basis kernel function kernel $k(\bmu, \bmu') = \exp(-||\bmu - \bmu'||^2 / 2) \in (0, 1]$ and a margin $t=0.5$.
Combining the above three objective functions with different weighting parameters $\alpha_{CH}$ and $\alpha_{CM}$, we train our proposed PVAE by maximizing
\begin{equation}
  \L (\theta, \phi; \X ) + \alpha_{CH} CH(\phi, \psi; \X) + \alpha_{CM} CM(\phi, \psi; \X, \tilde{\X}), \label{eq:pvae_obj}
\end{equation}
where a negative sample $\tilde{\X}$ is randomly sampled from the training set.

\section{Related Work}
Learning representations from multimodal data has gained significant interest in recent years~\cite{baltruvsaitis2018multimodal}.
Much work adopted some combination of text, speech, audio, image, or video~\cite{mansimov2016generating, leidal2017learning, harwath2017learning, arandjelovic2017objects, senocak2018learning, wang2018style}, and aimed to learn modality-invariant semantic representation from the combination.
For example, the authors in~\cite{harwath2016unsupervised} proposed a framework to learn concepts that are commonly described by an image and a parallel speech caption, but left out the speaker information and image-style information. 
Although such representations can be useful for certain tasks, such as pattern recognition or semantics-based retrieval, they cannot be applied to many other tasks that demand information appearing in one single modality, such as image generation or speaker verification. 
In contrast, we provide a unified PVAE framework for learning both the modality-invariant semantic information as well as the modality-dependent residual factors.

There have been many recent studies on learning disentangled representation using VAEs, where different sets of latent variables are sensitive to changes in different explanatory factors while being invariant to the rest of the factors~\cite{bengio2013representation}. 
One line of research adopts a simple graphical model that consists of only a single multi-dimensional latent variable, and aims for \textit{dimension-wise disentanglement}, relating each dimension to a different explanatory factor~\cite{Chen2016Infogan, higgins2017beta, kim2018disentangling, chen2018isolating}.
Disentanglement is often achieved by encouraging the representation distribution to be factorial.
However, although they are disentangled, such representations are not interpretable without manually inspecting the explanatory factors in the generated samples due to the exchangeability among dimensions of the latent variable.

Another line of research focuses on \textit{variable-wise disentanglement}, which encodes distinct aspects of data into separate latent variables~\cite{edwards2016towards, johnson2016composing, hsu2017unsupervised, narayanaswamy2017learning, li2018deep}.
These approaches incorporate the prior knowledge about the data generative process by designing graphical models with a fixed causal relationship among latent variables.
Representations learned via such methods are therefore interpretable without manual inspection.
One common trick of designing such graphical models is to tie one latent variable with the generation of multiple observed variables, such that this latent variable would encode the co-factors of the tied observed variables.
Our PVAE model shares the same view with Neural Statistician (NS)~\cite{edwards2016towards} and Factorized Hierarchical Variational Autoencoder (FHVAE)~\cite{hsu2017unsupervised} from this perspective, where we tie one latent variable to different modalities in a sample, instead of to instances in a dataset, or to segments in a sequence, as in the latter two models.

Our work is also related to the joint multimodal variational autoencoder (JMVAE)~\cite{suzuki2016joint}, which learns representations from multimodal data and allows bi-directional generation from one modality to another.
The authors assumed that the same set of explanatory factors is involved in the generative process of all modalities, and the generation of each modality is conditioned on the same latent variable.
Representations learned from the JMVAE are hence not disentangled.
Furthermore, the modalities that are considered in the JMVAE are images and their attribute labels, having a relatively deterministic mapping from the former to the latter.

\section{Experiments}

\subsection{Datasets, network architectures, and training}
For our experiments, we consider parallel spoken and written digits as our multimodal data.
The data have well-defined domain-invariant information (i.e. the digit identity), and contain rich modality-dependent factors, such as volume/timbre/duration for speech, and width/tilt/boldness for images.
Two spoken digit datasets are used in our experiments: TIDIGIT~\cite{leonard1993tidigits} and SecuVoice~\cite{martin2016secuvoice}. 
TIDIGIT contains over 17k broadband digit sequences in English spoken by 225 adult speakers.
We train a speech recognizer using Kaldi~\cite{povey2011kaldi} to separate sequences into individual digits.
SecuVoice contains over 3.5k sequences of isolated digits recorded with two smartphones, and we use recordings from the one with higher dynamic range.
An energy-based voice activity detection is applied to remove silence.
Speech is represented as a sequence of 80-dimensional Mel-scale filter bank coefficients (FBank), computed every 10ms.
The average digit duration is 0.37s and 0.76s for TIDIGIT and SecuVoice, respectively.
The MNIST dataset~\cite{lecun2010mnist} is used for written digits, and images are represented as $28 \times 28$ pixel matrices. 
To generate multimodal data, we randomly pair each spoken digit with a written digit of the same identity for each training epoch from the training partition of each dataset.
For clarity, we hereafter denote a multimodal sample as $(\x^a, \x^i)$, and their corresponding latent style variable as $(\z^a, \z^i)$, where the superscripts $a$ and $i$ refer to audio and image, respectively.

Here, we briefly describe the neural network architecture for each module. Illustrations and more details can be found in the Appendix.
For both audio and image decoders, $p(\x^a | \z^a, \z^s)$ and $p(\x^i | \z^i, \z^s)$, we assume unit variance and parameterize the mean with a neural network. 
The audio decoder is a one-layer long short-term memory network (LSTM)~\cite{hochreiter1997long} with $512$ cells that predicts one frame at a time and takes the same latent variables as input at each step.
The image decoder contains two feed forward layers with $512$ and $7 \times 7 \times 16$ units, followed by two transposed convolutional layers with $8$ and $1$ filters respectively, up-sampling by a factor of 2 each.
Regarding the inference networks, we apply a one-layer LSTM with $512$ cells to map variable-length speech to a fixed-dimensional vector, and apply two strided convolutional layers with $4$ and $8$ filters, followed by a fully-connected layer to map an image to a 512-dimensional vector.
These two modules are called pre-encoders.
The multimodal inference network feeds the concatenated audio and image pre-encoder outputs to a linear layer predicting the mean and variance of all latent variables, while the unimodal one takes its pre-encoder output as input to a linear layer predicting corresponding latent variables.
Parameters between unimodal and multimodal pre-encoders are not shared.
We set each latent variable $\z^s$, $\z^a$, $\z^i$ to be of 32 dimensions.
Adam~\cite{kingma2014adam} is used for training all parameters, $\theta$, $\phi$, and $\psi$, to maximize Eq.~\eqref{eq:pvae_obj}, with batch size $256$, initial learning rate $10^{-3}$, $\beta_1 = 0.95$, $\beta_2 = 0.999$, for 400 epochs.
If not otherwise mentioned, $\alpha_{CH}=0.1$ and $\alpha_{CM}=10$ are used.

\subsection{Baseline Models}
For the following experiments, we consider three baseline models: JMVAE-kl~\cite{suzuki2016joint} trained on the same multimodal data, and two VAEs~\cite{kingma2013auto} trained on only speech or images, denoted as VAE-sp and VAE-im, respectively.
To have a fair comparison, we let VAE-sp and VAE-im adopt the same encoder/decoder architectures as the corresponding PVAE modules.
As for JMVAE-kl, similar to PVAE, it also contains two generative networks, one multimodal inference network, and two unimodal inference networks;
however, there is only a single latent variable, inferred by all three inference networks, and conditioned by both generative networks.
We therefore construct a JMVAE-kl by setting all the network architectures to be the same as PVAE, and the latent variable dimension to be the sum of the dimensions of $\z^s$, $\z^a$, and $\z^i$.

\subsection{Existence of modality-dependent factors}
\begin{figure}[ht]
  \centering
  \includegraphics[width=\linewidth]{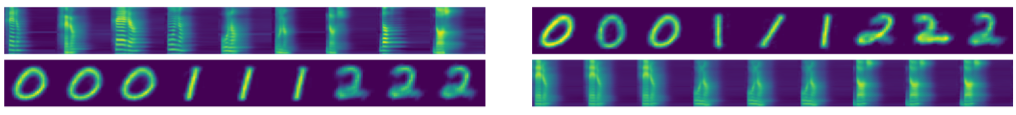}
  \caption{JMVAE-kl decoding results. Each column in the first and the second row on the left half of the figure is conditioned on the same latent variable inferred from a spoken digit using the unimodal inference network. On the right half of the figure, latent variables are inferred from written digits.}
  \label{fig:jmvae_inv}
\end{figure}

To confirm our assumption that there exist modality-dependent explanatory factors that only characterize the generation of a single modality, we demonstrate that changes of the latent variable in JMVAE-kl do not always result in the changes in both modalities.
Figures \ref{fig:jmvae_inv} illustrates the decoding results conditioned on the latent variables inferred from spoken digits (left) or written digits (right) using the corresponding unimodal inference network.
We can observe that representations inferred from different spoken samples are different, because they result in different spoken samples produced by the speech decoder (first row on the left); 
however, these representations always generate the same written samples if the digit identities are the same (second row on the left). 
The same results can also be observed when generating spoken digits from written digit-inferred representations.
These results imply that style in one modality is independent of style in the other, and therefore the guess of the style can be only as good as the prior.

\subsection{Evaluating disentanglement performance}
To study the performance of our model and training objectives on disentangling the shared semantic information from the modality-dependent style information, we need to show that:
\begin{enumerate*}[label=(\arabic*)]
	\item latent semantic variable $\z^s$ encodes only the digit identity information, and 
    \item latent style variables $\z^a$ and $\z^i$ contain style information, and combining one of them with $\z^s$ can reconstruct the corresponding modality.
\end{enumerate*}

We perform k-means clustering analysis on all three latent variables inferred from unimodal encoders to quantify the first criteria.
As there are only ten distinct semantic classes, an ideal semantic encoding would form ten non-overlapping clusters according to the digit identity. 
A cluster-size weighted average purity with respect to digit identity is therefore used as a metric, where cluster purity is defined as the proportion of instances within a cluster that have the same digit label as the majority digit label of that cluster.
Results of clustering with ten classes are listed in Table~\ref{tab:cluster}, where PVAE (no $CM$) refers to a PVAE model trained with $\alpha_{CM}=0$. 
For spoken digits, we can observe that samples with different digit identities are better separated in the latent semantic space of $\z^s$ learned by both PVAE models, than in the latent space learned with VAE-sp or JMVAE-kl.
In contrast, the latent style spaces of $\z^a$ and $\z^i$ contain little information about the digit identity, as desired.
Furthermore, when training with the proposed cross-modality semantic contrastiveness objective, PVAE disentanglement performance can be significantly improved, reaching accuracy of over 99\% in both modalities on both datasets.

\begin{figure}[ht]
  \centering
  \includegraphics[width=\linewidth]{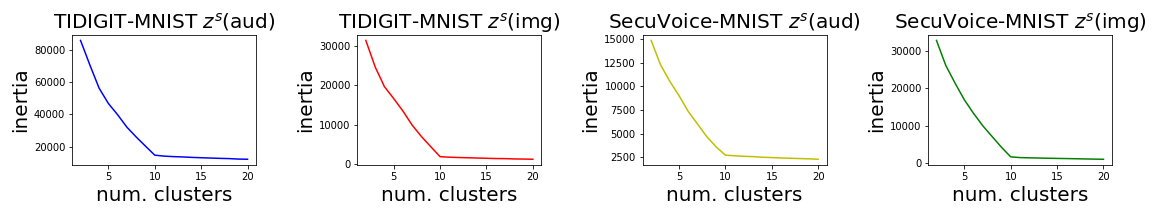}
  \caption{Inertia of k-means clustering with latent semantic variables versus the number of clusters.}
  \label{fig:cluster_var_to_n}
\end{figure}

In addition, we determine the ability of our model to automatically discover the number of semantic classes by plotting k-means clustering inertia of latent semantic variables with respect to the number of clusters, where inertia is defined as the sum of the squared Euclidean distance of each instance to the closest centroid.
Results in Figure~\ref{fig:cluster_var_to_n} show the same trend in all cases, where inertia decreases much slower after having ten clusters, indicating that PVAE learns to encode instances into the right number of classes in this space.
We also show t-SNE plots in the Appendix for visualization.

\begin{table}[ht]
  \caption{Results of weighted average purity based on digit identity. Clusters are obtained from 10-class k-means clustering on the latent variables. Multimodal datasets used are shown in the first row. The \textit{Speech} and \textit{Image} columns denote the modalities from which the latent variables are inferred. The symbol "-" indicates that inference from a modality is absent for that model. The \textit{Lat.} column shows the latent variable used for clustering, and \textit{Pur.} stands for weighted average purity.}
  \label{tab:cluster}
  \centering
  \begin{minipage}{.48\textwidth}
    \resizebox{\columnwidth}{!}{
      \begin{tabular}{lllll}
      \toprule
      \multicolumn{5}{c}{TIDIGIT-MNIST}                   \\
      \midrule
      \multirow{2}{*}{Model} & \multicolumn{2}{c}{speech} & \multicolumn{2}{c}{image} \\ 
      \cmidrule(l{2pt}r{2pt}){2-3} \cmidrule(l{2pt}r{2pt}){4-5}
      & Feat. & Pur.(\%) & Lat. & Pur.(\%) \\
      \midrule
      VAE-sp 						& $\z$ 		& 45.73 & - 		& -  \\ 
      VAE-im 						& - 		& - 	& $\z$		& 76.38  \\ 
      JMVAE-kl 						& $\z$ 		& 45.48 & $\z$ 		& 98.16 \\ 
      PVAE (no $CM$)				& $\z^{s}$ 	& 65.69 & $\z^{s}$	& 97.95 \\ 
      PVAE 							& $\z^{s}$ 	& \textbf{99.77} & $\z^{s}$	& \textbf{99.10} \\ 
      \midrule
      PVAE (no $CM$)  				& $\z^{a}$ 	& 30.88	& $\z^{i}$  & 21.73 \\
      PVAE  						& $\z^{a}$ 	& 24.79	& $\z^{i}$  & 20.63 \\
      \bottomrule
      \end{tabular}
    }
  \end{minipage}
  \begin{minipage}{.48\textwidth}
    \resizebox{\columnwidth}{!}{
      \begin{tabular}{lllll}
      \toprule
      \multicolumn{5}{c}{SecuVoice-MNIST}                   \\
      \midrule
      \multirow{2}{*}{Model} & \multicolumn{2}{c}{speech} & \multicolumn{2}{c}{image} \\ 
      \cmidrule(l{2pt}r{2pt}){2-3} \cmidrule(l{2pt}r{2pt}){4-5}
      & Feat. & Pur.(\%) & Lat. & Pur.(\%) \\
      \midrule
      VAE-sp 						& $\z$ 		& 63.73 & - 		& -  \\ 
      VAE-im 						& - 		& - 	& $\z$		& 76.38  \\ 
      JMVAE-kl 						& $\z$ 		& 64.43 & $\z$ 		& 98.46 \\
      PVAE (no $CM$)				& $\z^{s}$ 	& 77.80 & $\z^{s}$	& 98.43 \\ 
      PVAE 							& $\z^{s}$ 	& \textbf{99.30} & $\z^{s}$	& \textbf{99.02} \\ 
      \midrule
      PVAE (no $CM$)				& $\z^{a}$ 	& 33.69 & $\z^{i}$	& 22.43 \\
      PVAE  						& $\z^{a}$ 	& 25.78	& $\z^{i}$  & 21.81 \\
      \bottomrule
      \end{tabular}
    }
  \end{minipage}
\end{table}

\begin{figure}[ht]
  \centering
  \includegraphics[width=\linewidth]{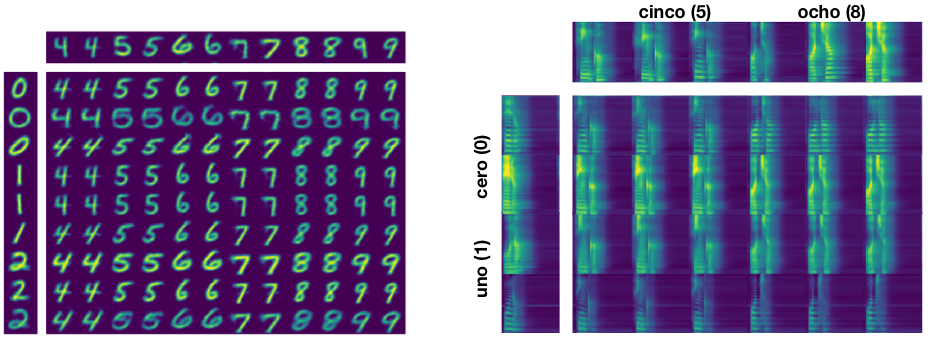}
  \caption{(left) Written digits generated by varying latent variables. The top row shows samples for inferring $\z^s$, and the leftmost column shows samples for inferring $\z^i$. In the lower right block, each column conditions on the same $\z^s$ and each row conditions on the same $\z^i$. (right) Spanish spoken digits generated by varying latent variables with a similar approach. We show more samples in the Appendix.}
  \label{fig:lat_recomb}
\end{figure}

To demonstrate that the $\z^s$ is invariant to style change, and $\z^a$ and $\z^i$ encode residual style information, we infer these latent variables from a set of instances and recombine them to generate new instances.
Results are shown in Figure~\ref{fig:lat_recomb}.
For both written and spoken digits, we can observe that instances in each row are consistent in style,\footnote{For FBank images of spoken digits, the most salient style attributes are duration, volume, and pitch, which correspond to width, intensity, and horizontal stripe spacing. In Figure~\ref{fig:lat_recomb} (right), row one has the highest pitch, row four has the shortest duration and lowest volume, and row two has the highest volume.} and are visually identical between columns when conditioned on $\z^s$ of the same digit.
This qualitative analysis verifies the ability of our PVAE model to factorize semantic information and style information.

\subsection{Controlled cross-modal generation}
With semantic and style factors separated into different latent variables, in addition to modifying the speaking style of a spoken digit, or transferring the writing style of a written digit, we can also apply the PVAE model for controlled cross-modal generation.
For example, a written digit can be converted to a spoken one, with the spoken style specified by a reference spoken digit, regardless of its identity; 
likewise, a spoken digit can also be converted to a written one with a specified style.

\begin{figure}[ht]
  \centering
  \includegraphics[width=\linewidth]{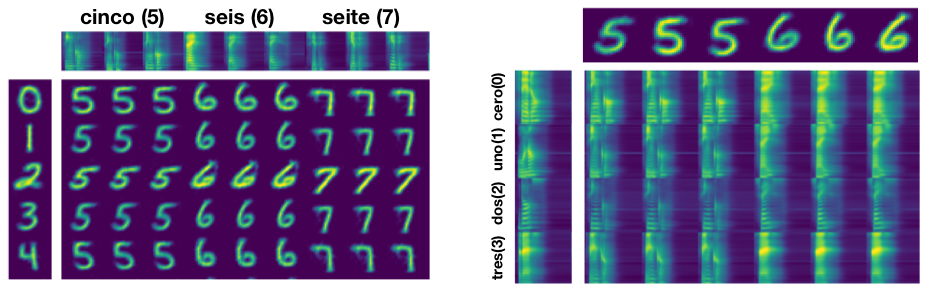}
  \caption{(left) Generating written digits from spoken digits with the specified writing styles. (right) Generating spoken digits from written digits with the specified speaking styles. Figures are arranged in a similar fashion as in Figure~\ref{fig:lat_recomb}, with $\z^s$ inferred from the top row, and $\z^i/\z^a$ from the leftmost column. More samples are in the Appendix.}
  \label{fig:cm_recomb}
\end{figure}

For the former scenario, we infer the latent semantic variable $\z^s$ from a written digit, and the latent style variable $\z^a$ from a reference spoken digit.
Conditioning on this pair $\z^s$ and $\z^a$, we generate a new spoken digit using the speech generative network.
By swapping the modalities, we arrive at the latter case and generate new written digits using the image generative network.
Figure~\ref{fig:cm_recomb} shows conversion from spoken digits in Spanish to written digits on the left, and from written digits to spoken digits in Spanish on the right.
Both directions of generation succeed in referencing the specified style while maintaining the correct digit identity.

\section{Conclusion and Future Work}
We investigate a learning scenario with parallel multimodal sensory data that resembles the form of information humans perceive during their learning process. 
A novel PVAE model is proposed for learning disentangled representation from such data, separating modality-invariant semantic information from modality-dependent style information.
In addition to the usual variational lower bound, we also propose two objectives for learning unimodal inference networks, as well as promoting disentanglement.
Experimental results demonstrate the success of our proposed model and objectives in discovering the latent semantic information without explicitly assigning one-hot labels, and separating it and the rest explanatory factors into different latent variables.

Our proposed framework is general and can be easily applied to learning from other parallel multimodal data, such as audio-visual speech data with audio recordings and videos of the person talking~\cite{dupont2000audio, mroueh2015deep}, or parallel audios and videos of instrument playing~\cite{gemmeke2017audio, arandjelovic2017objects}.
Extending from our proposed partitioned generative process, we also plan to investigate a more sophisticated setting, where multiple concepts are presented in a multimodal sample, while some may only be observed in partial modalities.

\medskip
\clearpage

{
\small 
\bibliographystyle{plain}
\bibliography{paper_supp.bib}
}

\clearpage

\section*{A. Derivation of PVAE variational lower bound}
The variational lower bound on the log marginal likelihood of a multimodal sample $\X = \{ \x^m \}_{m=1}^M$ can be derived as follows:
\begin{align}
  \log p(\X) =& \log \E_{ q( \{ \z^m \}_{m=1}^M, \z^s | \{ \x^m \}_{m=1}^M ) } \big[ \dfrac{ p( \{ \x^m \}_{m=1}^M, \{ \z^m \}_{m=1}^M, \z^s) }{ q( \{ \z^m \}_{m=1}^M, \z^s | \{ \x^m \}_{m=1}^M ) } \big]  \\
  \geq& \E_{ q( \{ \z^m \}_{m=1}^M, \z^s | \{ \x^m \}_{m=1}^M ) } \big[ \log \dfrac{ p( \{ \x^m \}_{m=1}^M, \{ \z^m \}_{m=1}^M, \z^s) }{ q( \{ \z^m \}_{m=1}^M, \z^s | \{ \x^m \}_{m=1}^M ) } \big] \label{eq:js} \\
  =& \E_{ q( \{ \z^m \}_{m=1}^M, \z^s | \{ \x^m \}_{m=1}^M ) } \big[ \log \dfrac{p(\z^s) \prod_{m=1}^M p(\z^m) p( \x^m | \z^m, \z^s)}{q(\z^s | \{ \x^m \}_{m=1}^M ) \prod_{m=1}^M q(\z^m | \{ \x^m \}_{m=1}^M )} \big] \label{eq:expand} \\
  =& \sum_{m=1}^M \E_{ q( \{ \z^m \}_{m=1}^M, \z^s | \{ \x^m \}_{m=1}^M ) } \big[ \log p( \x^m | \z^m, \z^s) \big] \nonumber \\
  &- \sum_{m=1}^M \E_{ q( \{ \z^m \}_{m=1}^M, \z^s | \{ \x^m \}_{m=1}^M ) } \big[ \log \dfrac{ q(\z^m | \{ \x^m \}_{m=1}^M )}{ p(\z^m) } \big] \nonumber \\
  &- \E_{ q( \{ \z^m \}_{m=1}^M, \z^s | \{ \x^m \}_{m=1}^M ) } \big[ \log \dfrac{q(\z^s | \{ \x^m \}_{m=1}^M )}{p(\z^s)} \big] \label{eq:reorg} \\
  =& \sum_{m=1}^M \E_{ q( \{ \z^m \}_{m=1}^M, \z^s | \{ \x^m \}_{m=1}^M ) } \big[ \log p( \x^m | \z^m, \z^s) \big] \nonumber \\
  &- \sum_{m=1}^M \E_{ q(\z^m | \{ \x^m \}_{m=1}^M ) } \big[ \log \dfrac{ q(\z^m | \{ \x^m \}_{m=1}^M ) }{ p(\z^m) } \big] \nonumber \\
  &- \E_{ q(\z^s | \{ \x^m \}_{m=1}^M ) } \big[ \log \dfrac{q(\z^s | \{ \x^m \}_{m=1}^M )}{p(\z^s)} \big] \label{eq:rm_ind} \\
  =& \sum_{m=1}^M \E_{ q( \{ \z^m \}_{m=1}^M, \z^s | \{ \x^m \}_{m=1}^M ) } \big[ \log p( \x^m | \z^m, \z^s) \big] \nonumber \\
  &- \sum_{m=1}^M D_{KL} ( q(\z^m | \{ \x^m \}_{m=1}^M ) || p(\z^m) ) \nonumber \\
  &- D_{KL} ( q(\z^s | \{ \x^m \}_{m=1}^M ) || p(\z^s) ) \label{eq:rewrite} \\
  =& \L (p, q; \X ),
\end{align}
where the Jensen inequality is applied to derive Eq.~\eqref{eq:js}, which is then expanded to Eq.~\eqref{eq:expand} based on the factorization assumptions of the proposed multimodal generative and inference networks.
We reorganize Eq.~\eqref{eq:expand} to obtain Eq.~\eqref{eq:reorg}, and by applying the factorization assumption again and marginalizing over latent variables that do not appear within the expectations, the last two terms in Eq.~\eqref{eq:rm_ind} can be derived from the last two terms in Eq.~\eqref{eq:reorg}. 
Lastly, the last two terms in Eq.~\eqref{eq:rm_ind} can be re-written using the KL-divergence notation, and we arrive at our variational lower bound in Eq.~\eqref{eq:rewrite}.

\section*{B. PVAE neural network architectures}
\begin{figure}[ht]
  \centering
  \begin{minipage}{.6\textwidth}
    \centering
    \includegraphics[width=\linewidth]{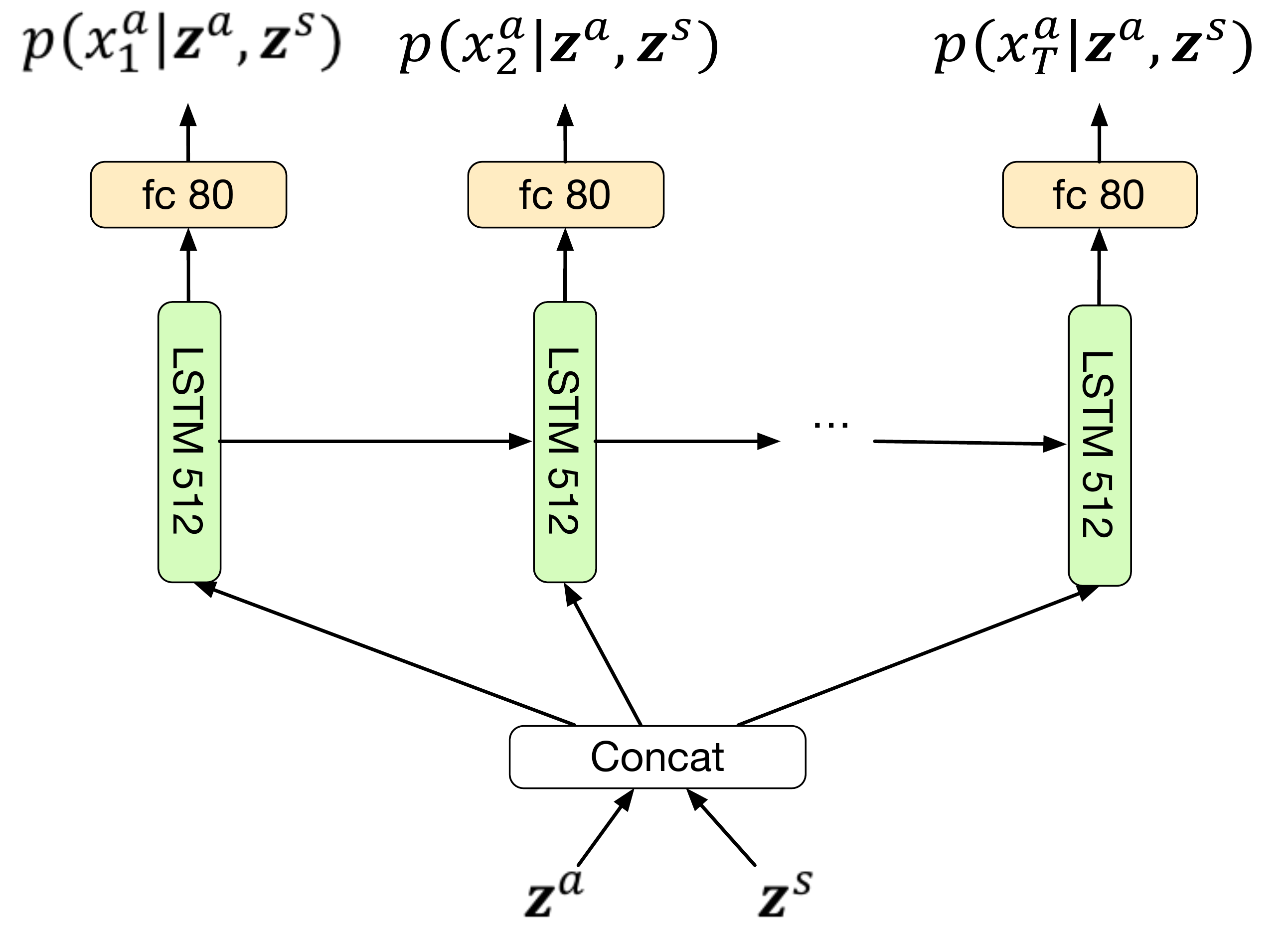}
    \caption{PVAE audio generative network.}
    \label{fig:pvae_aud_gen}
  \end{minipage}
  \hspace{1cm}
  \begin{minipage}{.3\textwidth}
    \centering
    \includegraphics[width=\linewidth]{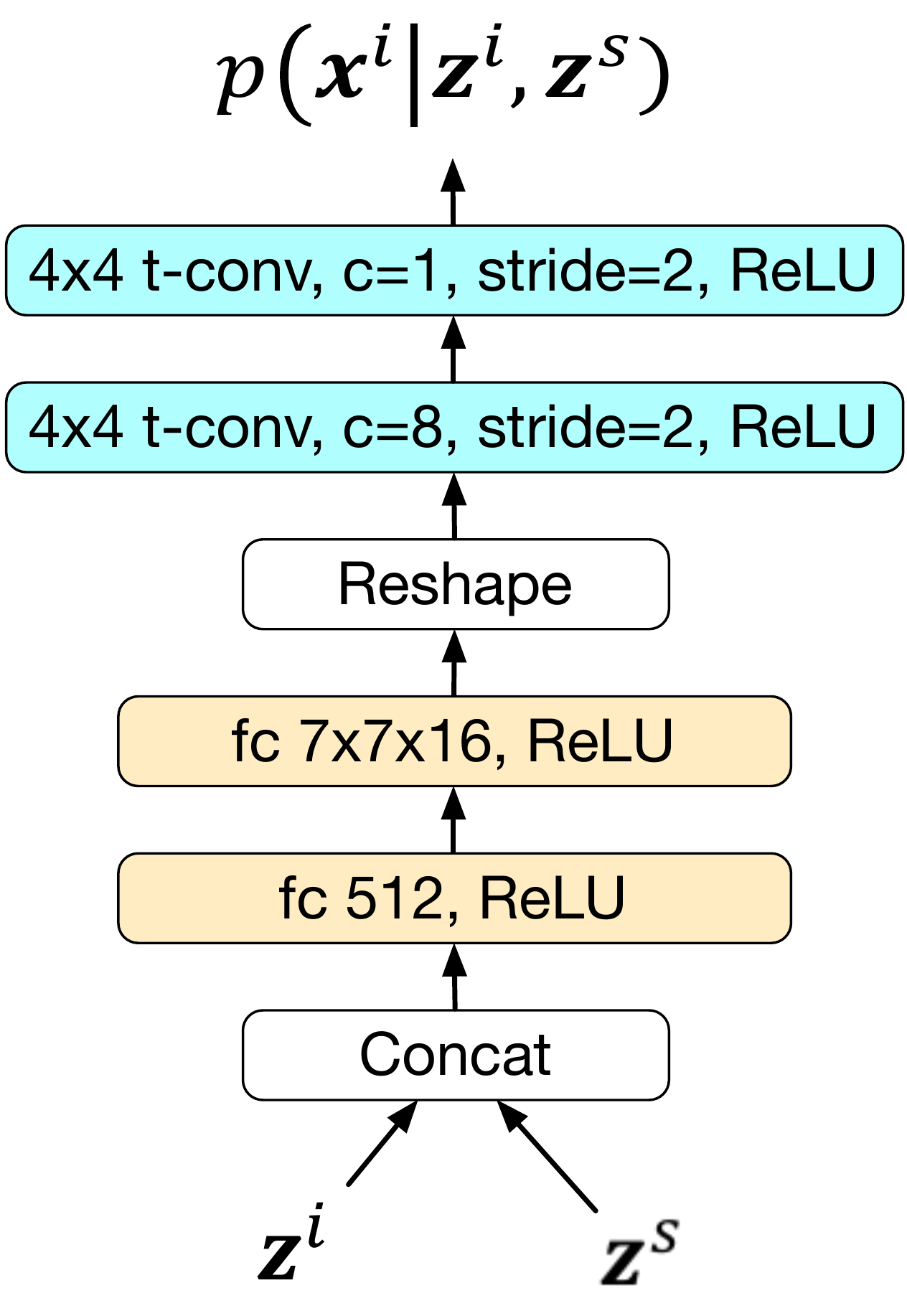}
    \caption{PVAE image generative network.}
    \label{fig:pvae_img_gen}
  \end{minipage}
\end{figure}

In this section, we explain in detail the network architecture of each module in PVAE and illustrate them graphically. 
Let $\x^i$ denote an image sample, and $\x^a = \{ x^a_1, x^a_2, \cdots, x^a_T \}$ denote a speech sample of $T$ frames.
Figures \ref{fig:pvae_aud_gen} and \ref{fig:pvae_img_gen} show the audio generative network and the image generative network, respectively.
For the audio generative network, the latent semantic variable $\z^s$ and the latent style variable for audio $\z^a$ are concatenated, and taken as input at each step to a one-layer LSTM with 512 cells.
The output of the LSTM at each step is passed to a fully-connected layer, which is shared among time steps, to predict the mean of the conditional distribution $p(x_t^a | \z^a, \z^s)$ for the corresponding frame.
For the image generative network, the latent semantic variable $\z^s$ and the latent style variable for image $\z^i$ are concatenated before feeding into the subsequent fully-connected layers, with 512 and 7$\times$7$\times$16 hidden units, and rectified linear units as activations, respectively.
The output from the last fully-connected layer is reshaped to a 7$\times$7$\times$16 tensor, and then passed to two transposed convolutional layers. 
The first convolutional layer has a 4$\times$4 kernel of 8 channels, up-sampling its input by a factor of two; 
the second one has a 4$\times$4 kernel of 1 channel, up-sampling its input by a factor of two to predict the mean of the conditional distribution $p(\x^i | \z^a, \z^s)$ .

\begin{figure}[ht]
  \centering
  \includegraphics[width=.7\linewidth]{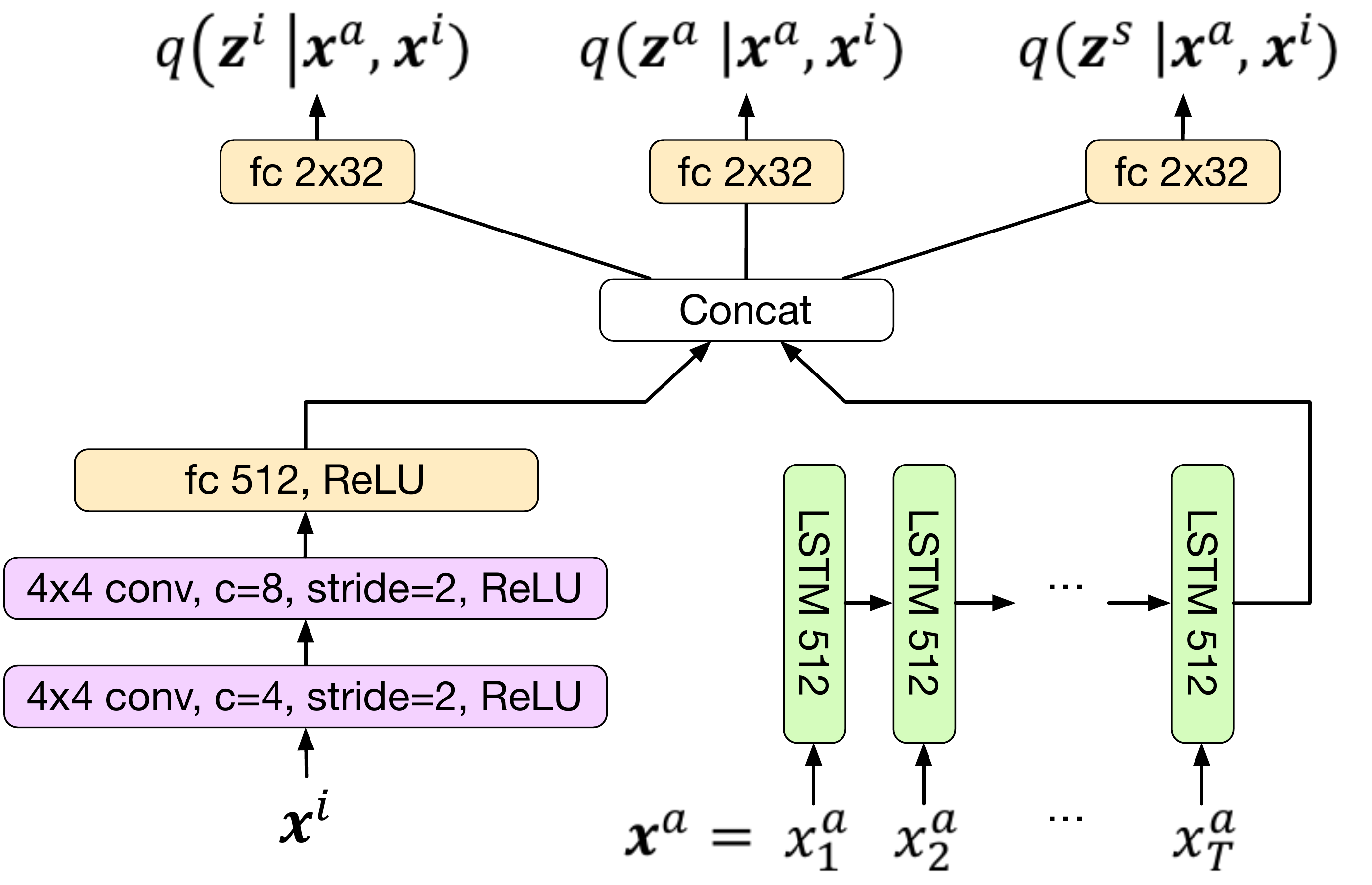}
  \caption{PVAE multimodal inference network.}
  \label{fig:pvae_multi_inf}
\end{figure}

\begin{figure}[ht]
  \centering
  \begin{minipage}{.4\textwidth}
    \centering
    \includegraphics[width=\linewidth]{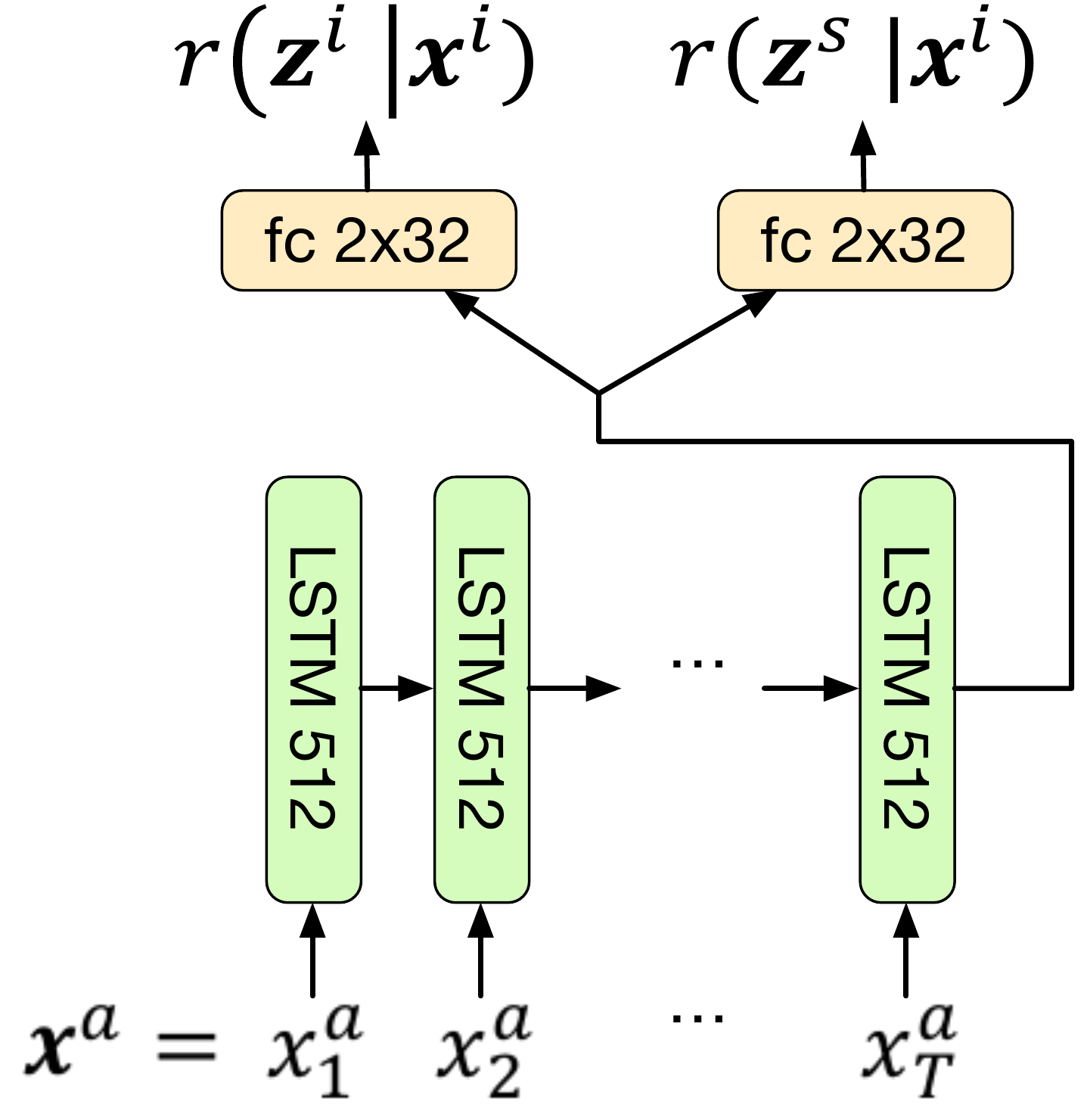}
    \caption{PVAE audio inference network.}
    \label{fig:pvae_aud_inf}
  \end{minipage}
  \hspace{1cm}
  \begin{minipage}{.3\textwidth}
    \centering
    \includegraphics[width=\linewidth]{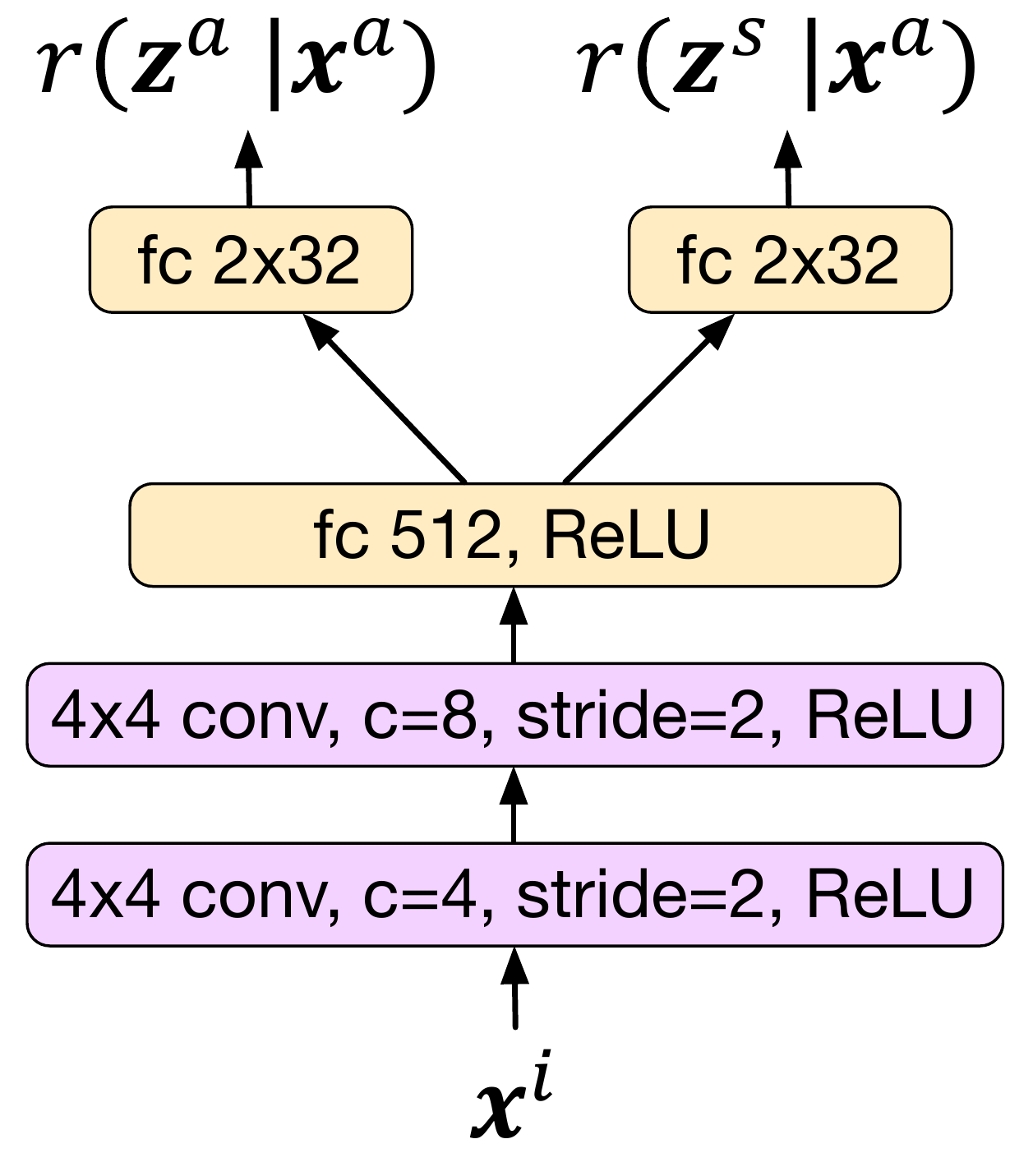}
    \caption{PVAE image inference network.}
    \label{fig:pvae_img_inf}
  \end{minipage}
\end{figure}

Figure~\ref{fig:pvae_multi_inf} illustrates the multimodal inference network, and Figures~\ref{fig:pvae_aud_inf} and \ref{fig:pvae_img_inf} show the unimodal audio and image inference network, respectively.
For both the multimodal and unimodal inference networks, we use similar neural network modules, referred to as the \textit{pre-encoders}, to encode each modality into a fixed-dimensional vector.

The audio pre-encoder is a one-layer LSTM with 512 cells that maps a variable-length sequence into a 512-dimensional vector.
The image pre-encoder consists of two 4$\times$4 down-sampling convolutional layers with stride$=2$ and number of channels=$\{4, 8\}$, followed by a fully-connected layer of 512 hidden units, to encode an image into a 512-dimensional vector.
For the multimodal inference network, outputs from the pre-encoders are concatenated and taken as input to three different fully-connected layers that predict the mean and the variance vectors of $q(\z^i | \x^a, \x^i)$, $q(\z^a | \x^a, \x^i)$, and $q(\z^s | \x^a, \x^i)$, respectively.
For the unimodal inference network, output from the pre-encoder is passed to two different fully-connected layers, predicting the mean and variance vectors of $r(\z^s | \x^a)$ and $r(\z^a | \x^a)$ if inferring from audio, or $r(\z^s | \x^i)$ and $r(\z^i | \x^i)$ if inferring from images.
Note that the parameters of the pre-encoders are not shared among the multimodal and the unimodal inference networks.

\section*{C. Visualization of PVAE latent spaces}
The ability of our proposed PVAE model to separate semantic and style information into different latent variables is quantitatively verified with k-means clustering analysis in Section 4.
In this section, we provide additional qualitative analysis by visualizing each PVAE latent variable using t-Distributed Stochastic Neighbor Embedding (t-SNE)~\cite{maaten2008visualizing}, a dimension reduction technique for embedding high-dimensional datasets into a low-dimensional space for visualization.

\begin{figure}[ht]
  \centering
  \begin{minipage}{.44\textwidth}
    \centering
    \includegraphics[width=\linewidth]{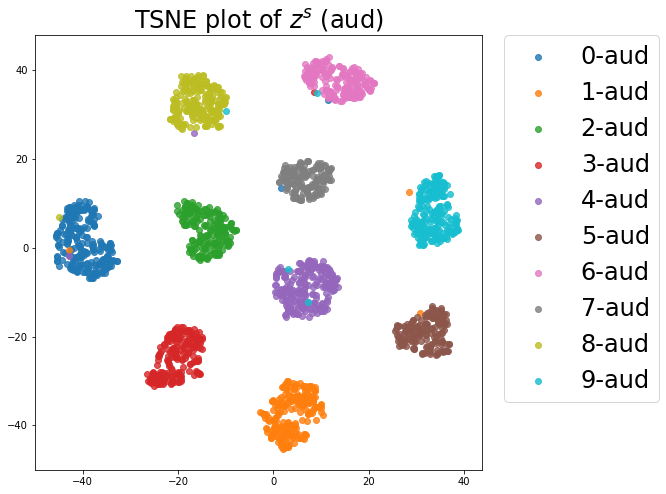}
    \caption{T-SNE plot of $\z^s$ inferred from unimodal speech samples. The samples form ten clusters based on the digit identity.}
    \label{fig:tsne_zs_aud}
  \end{minipage}
  \hspace{.7cm}
  \begin{minipage}{.44\textwidth}
    \centering
    \includegraphics[width=\linewidth]{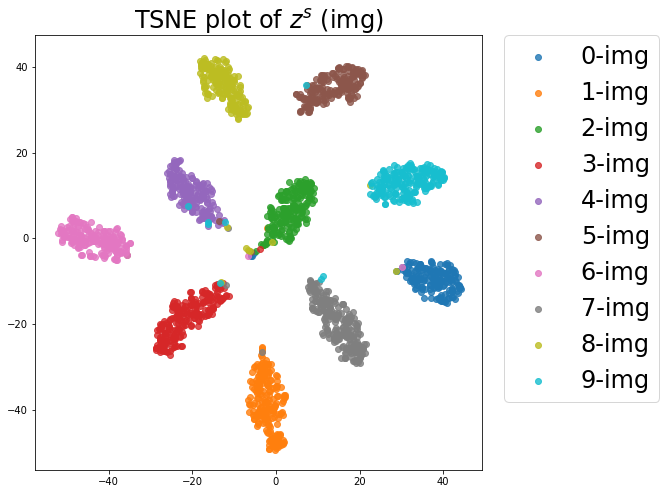}
    \caption{T-SNE plot of $\z^s$ inferred from unimodal image samples. The samples form ten clusters based on the digit identity.}
    \label{fig:tsne_zs_img}
  \end{minipage}
\end{figure}

We present the results of a PVAE model trained on the SecuVoice-MNIST dataset.
Figures \ref{fig:tsne_zs_aud} and \ref{fig:tsne_zs_img} show the projected embeddings of the latent semantic variables $\z^s$, inferred from unimodal speech and image samples, respectively.
We can clearly observe ten disconnected clusters in each plot, consistent with the inertia results from k-means clustering with different numbers of clusters.

\begin{figure}[ht]
  \centering
  \includegraphics[width=.7\linewidth]{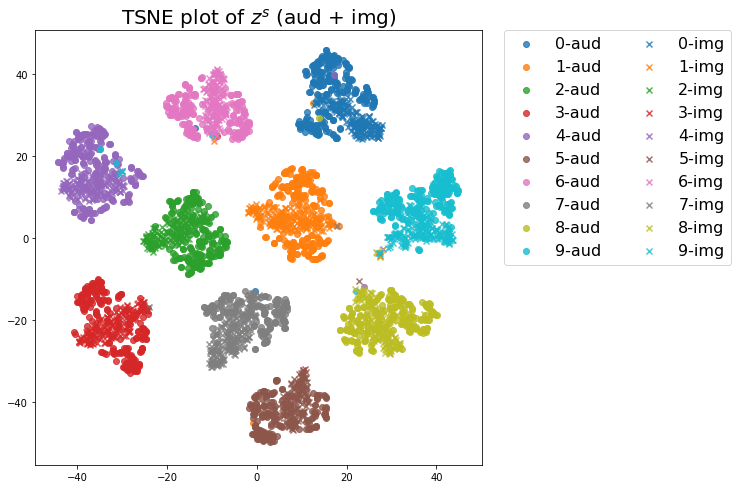}
  \caption{T-SNE plot of $\z^s$ inferred from unimodal speech samples and unimodal image samples. Images samples are marked with "x", and speech samples are marked with solid "o". The samples form ten clusters based on the digit identity, regardless of the modality they are inferred from}
  \label{fig:tsne_zs_aud_img}
\end{figure}

In order to examine whether samples of the same digit identity from different modalities are encoded in the same neighborhood, we further pool together the inferred latent semantic variables from each modality, and plot their embeddings with different markers in Figure~\ref{fig:tsne_zs_aud_img}, color-coded by the digit identity.
Similarly to the previous results, the pooled latent semantic variables also form ten clusters, where samples of the same digit identity share a cluster, regardless of the modality they are inferred from.

\begin{figure}[ht]
  \centering
  \begin{minipage}{.44\textwidth}
    \centering
    \includegraphics[width=\linewidth]{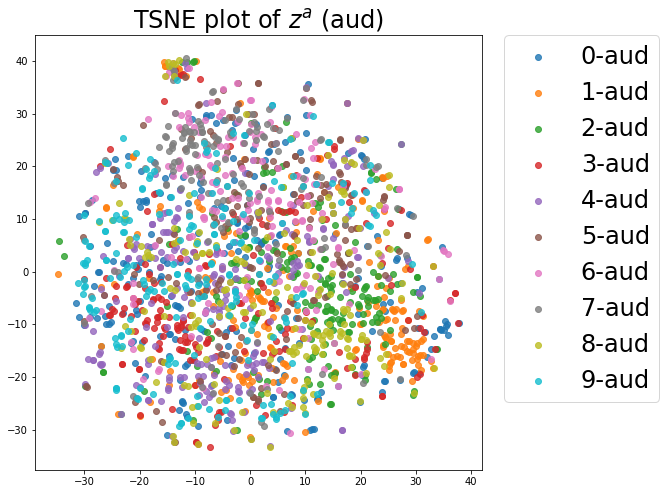}
    \caption{T-SNE plot of $\z^a$ inferred from unimodal speech samples. No visible clusters are formed based on the digit identity.}
    \label{fig:tsne_za_aud}
  \end{minipage}
  \hspace{.7cm}
  \begin{minipage}{.44\textwidth}
    \centering
    \includegraphics[width=\linewidth]{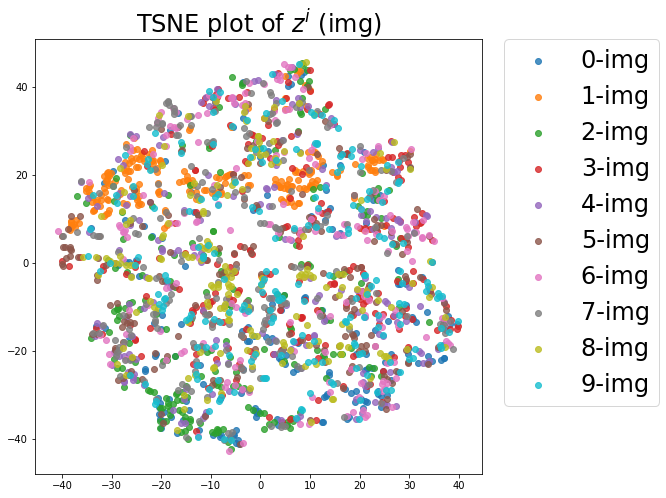}
    \caption{T-SNE plot of $\z^i$ inferred from unimodal image samples. No visible clusters are formed based on the digit identity.}
    \label{fig:tsne_zi_img}
  \end{minipage}
\end{figure}

We next plot the latent style variables, $\z^a$ and $\z^i$, inferred from audio and image samples, respectively, in Figures \ref{fig:tsne_za_aud} and \ref{fig:tsne_zi_img}.
It can be observed that the distributions of the latent style variables are much less correlated to the digit identity, compared to those of the latent semantic variables.
These results are consistent with the quantitative analysis based on k-means clustering, both of which suggest that our proposed PVAE model is capable of separating digit identity and style into different latent variables.

\section*{D. Additional examples of within-modality style transformation}
In Section 4, we demonstrate on a subset of digit combinations that PVAE can transfer the style of a written or spoken digit without changing its identity, by fixing the latent semantic variable $\z^s$ and altering only the latent style variable $\z^a$ or $\z^i$.
Here we provide additional examples of conditioning the generation on latent semantic variables and latent style variables inferred from all ten digits, in order to further verify the effectiveness.

\begin{figure}[ht]
  \centering
  \includegraphics[width=\linewidth]{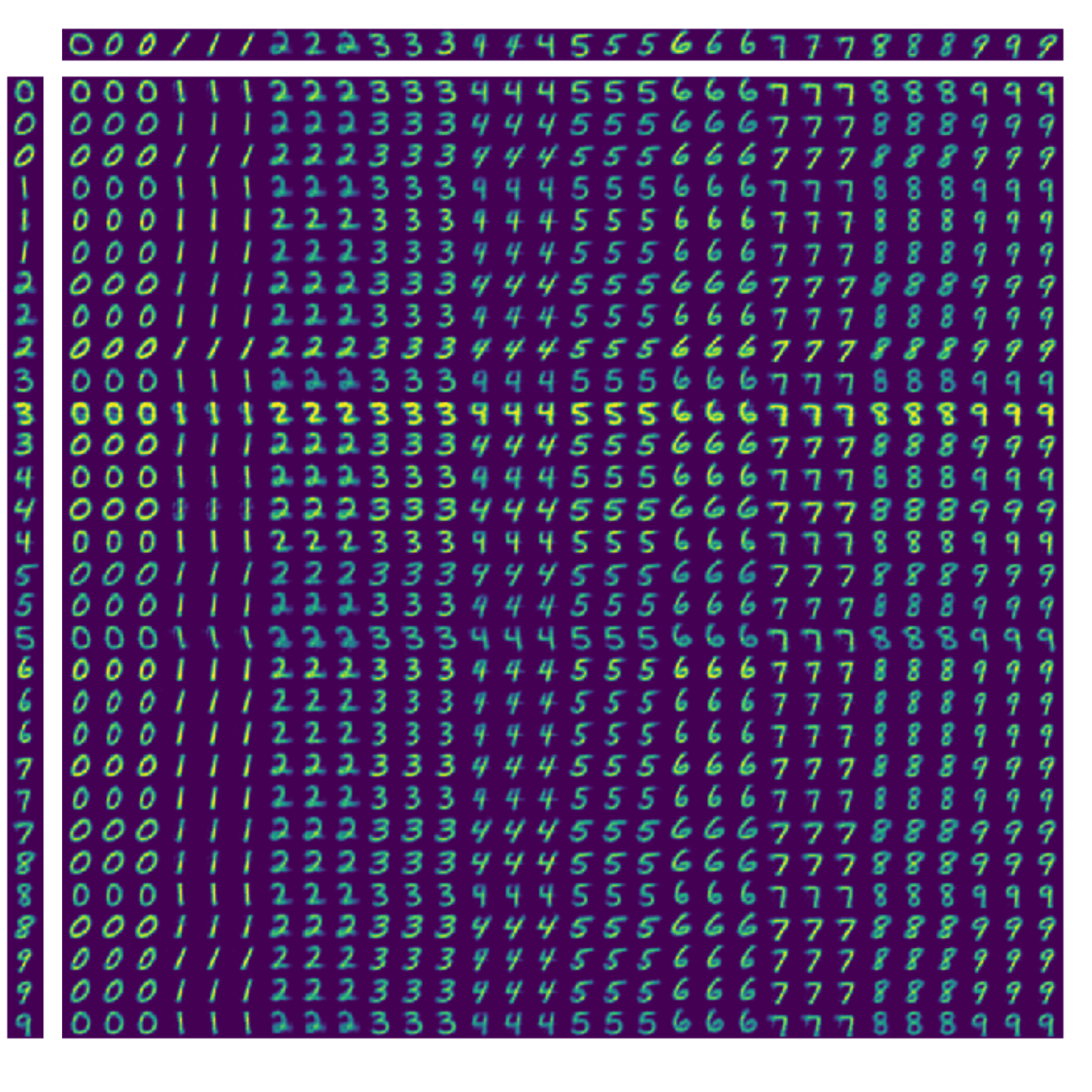}
  \caption{Written digit style transformation. Samples within each column are conditioned on the same latent semantic variable $\z^s$ inferred from the image in the top row, and samples within each row are conditioned on the same latent image style variable $\z^i$ inferred from the image in the leftmost column. The digit identities are consistent within each column, and the writing styles are consistent within each row.}
  \label{fig:within_img}
\end{figure}

Figure~\ref{fig:within_img} shows the style transformation results for written digits, where three samples for each digit on the top row are used to infer the latent semantic variables $\z^s$, and three samples for each digit on the leftmost column are used to infer the latent style variables $\z^i$.
We can observe that the digit identities are preserved within each column, and the style attributes, such as tilting, width, and stroke boldness, are consistent within each row.

\begin{figure}[ht]
  \centering
  \includegraphics[width=\linewidth]{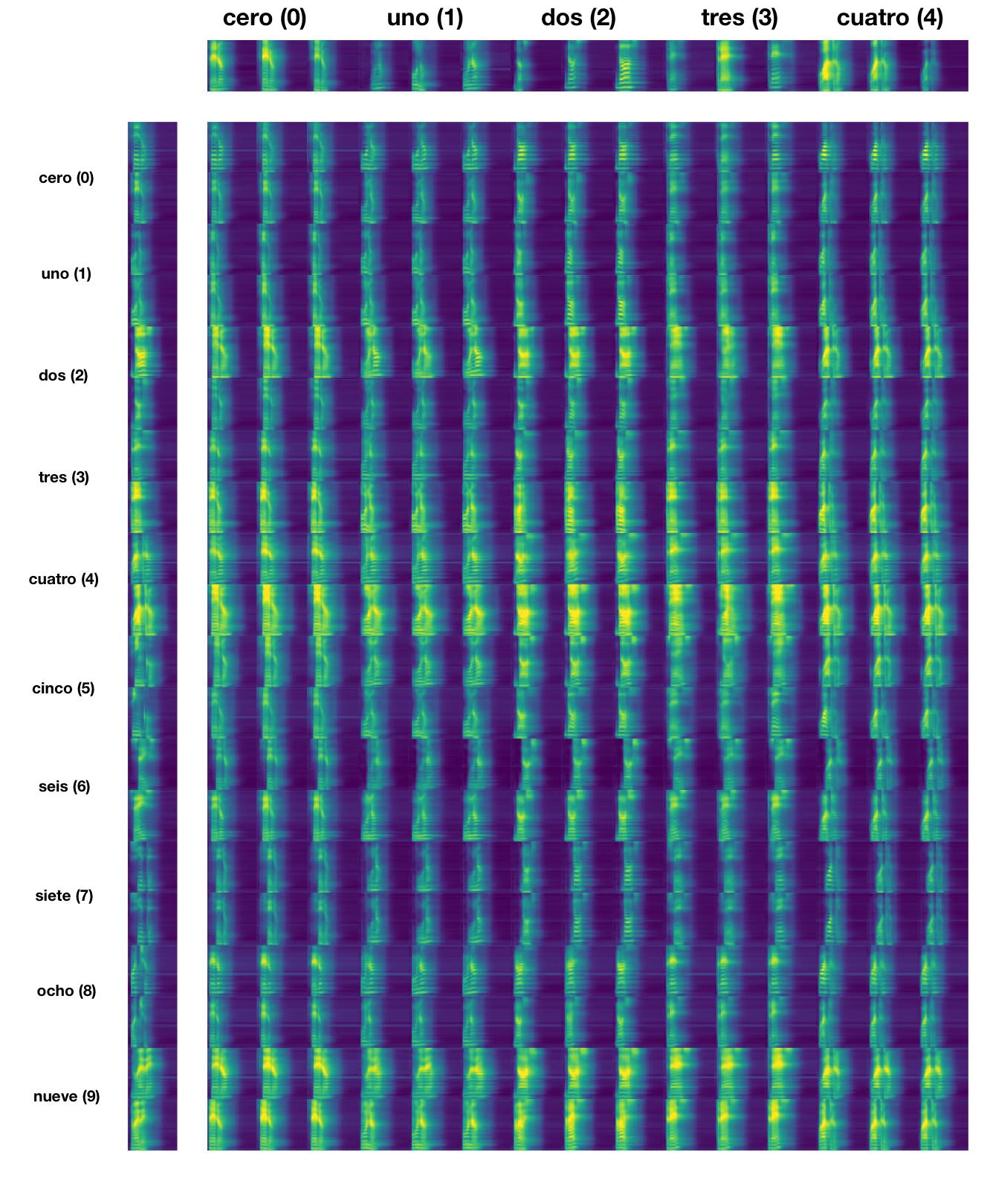}
  \caption{Spanish Spoken digit style transformation from 0 to 4. Samples within each column are conditioned on the same latent semantic variable $\z^s$ inferred from the audio in the top row, and samples within each row are conditioned on the same latent audio style variable $\z^a$ inferred from the audio in the leftmost column. The digit identities are consistent within each column, and the speaking styles are consistent within each row.}
  \label{fig:within_aud1}
\end{figure}

\begin{figure}[ht]
  \centering
  \includegraphics[width=\linewidth]{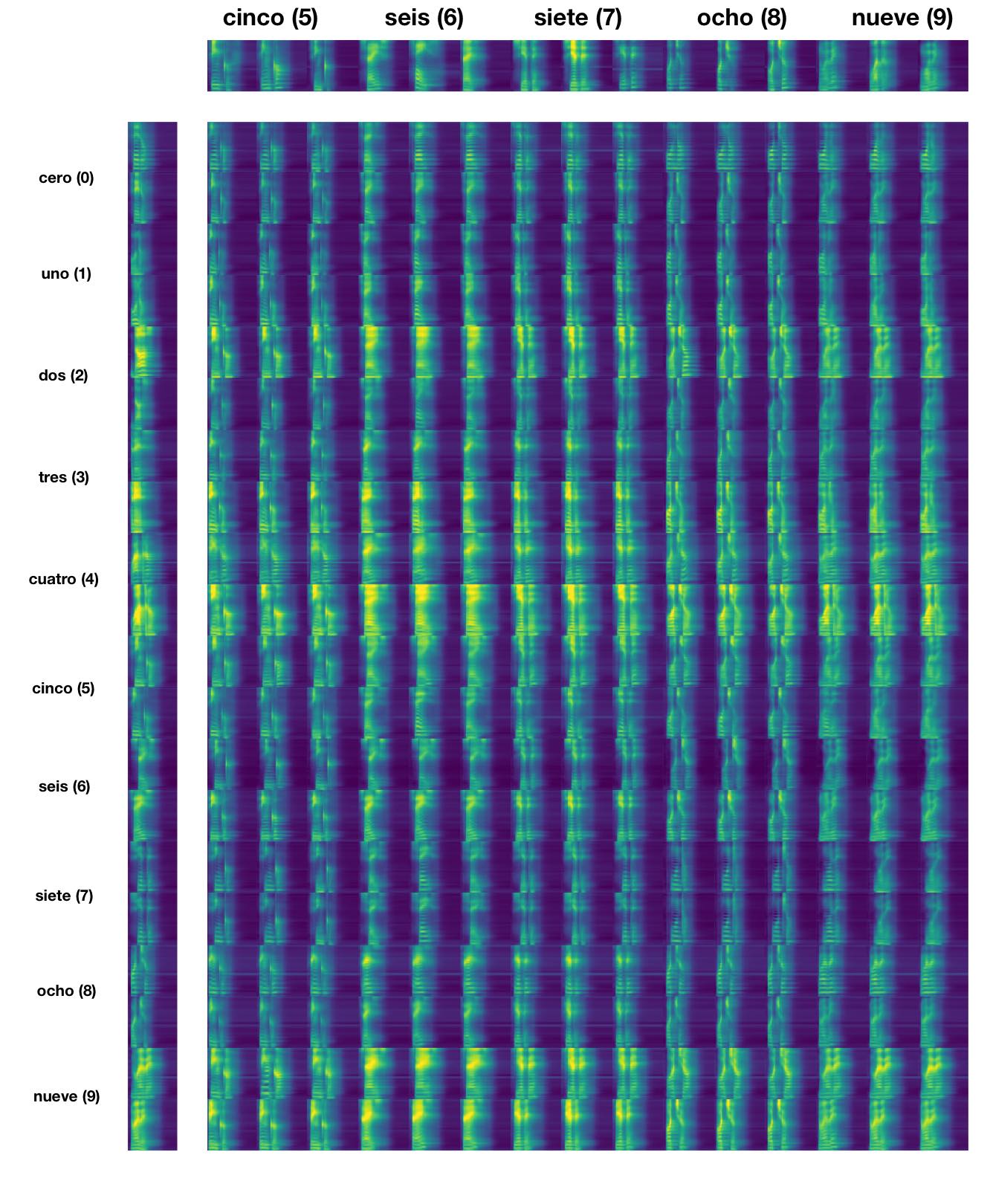}
  \caption{Spanish Spoken digit style transformation from 5 to 9. Samples within each column are conditioned on the same latent semantic variable $\z^s$ inferred from the audio in the top row, and samples within each row are conditioned on the same latent audio style variable $\z^a$ inferred from the audio in the leftmost column. The digit identities are consistent within each column, and the speaking styles are consistent within each row.}
  \label{fig:within_aud2}
\end{figure}

Figures~\ref{fig:within_aud1} and \ref{fig:within_aud2} show the style transformation results for spoken digits in Spanish, where three samples for each digit on the top row are used to infer the latent semantic variables $\z^s$, and two samples for each digit on the leftmost column are used to infer the latent style variables $\z^a$.
Again, the digit identities are preserved within each column, which can be identified based on the spectral contour and temporal position of formants, as well as the relative position of the formants.
The speaking style attributes, such as volume, pitch, duration, and starting time offset, are also consistent within each row.

\clearpage

\section*{E. Additional examples of controlled cross-modality generation}
In Section 5, we demonstrate that PVAE can generate a written digit from a spoken digit in Spanish with a specified writing style, or generate a spoken digit in Spanish from a written digit with a specified speaking style.
In this section, we present additional examples using two models that are trained on TIDIGIT-MNIST and SecuVoice-MNIST datasets respectively, and apply such conditional generation to all digit combinations from which the latent semantic variables and the latent style variables are inferred.

\begin{figure}[ht]
  \centering
  \includegraphics[width=\linewidth]{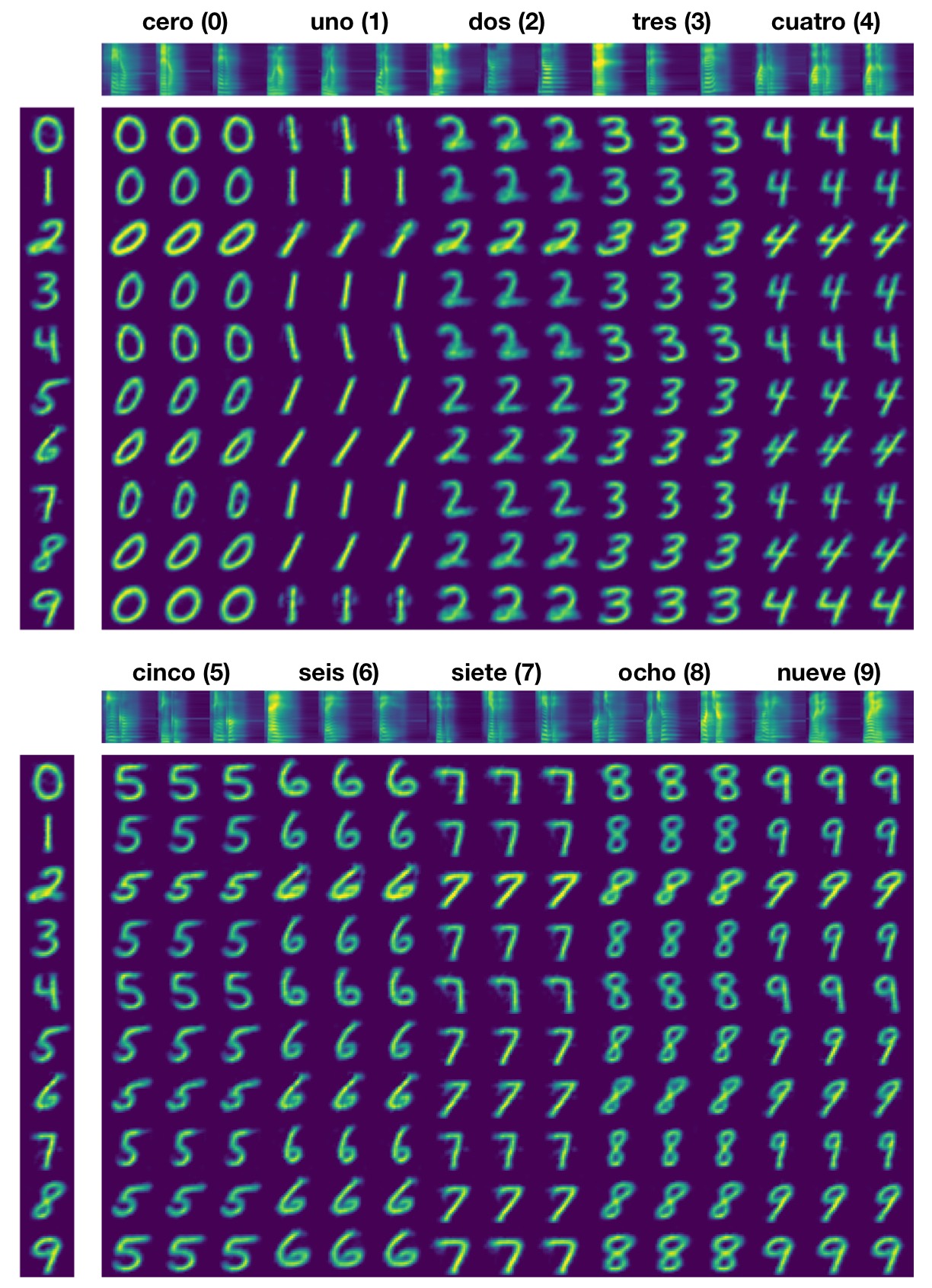}
  \caption{Generating written digits from spoken digits in Spanish. Samples within each column are conditioned on the same latent semantic variable $\z^s$ inferred from the audio in the top row, and samples within each row are conditioned on the same latent image style variable $\z^i$ inferred from the image in the leftmost column. The digit identities are consistent within each column, and the writing styles are consistent within each row.}
  \label{fig:spn_aud2img}
\end{figure}

\begin{figure}[ht]
  \centering
  \includegraphics[width=.9\linewidth]{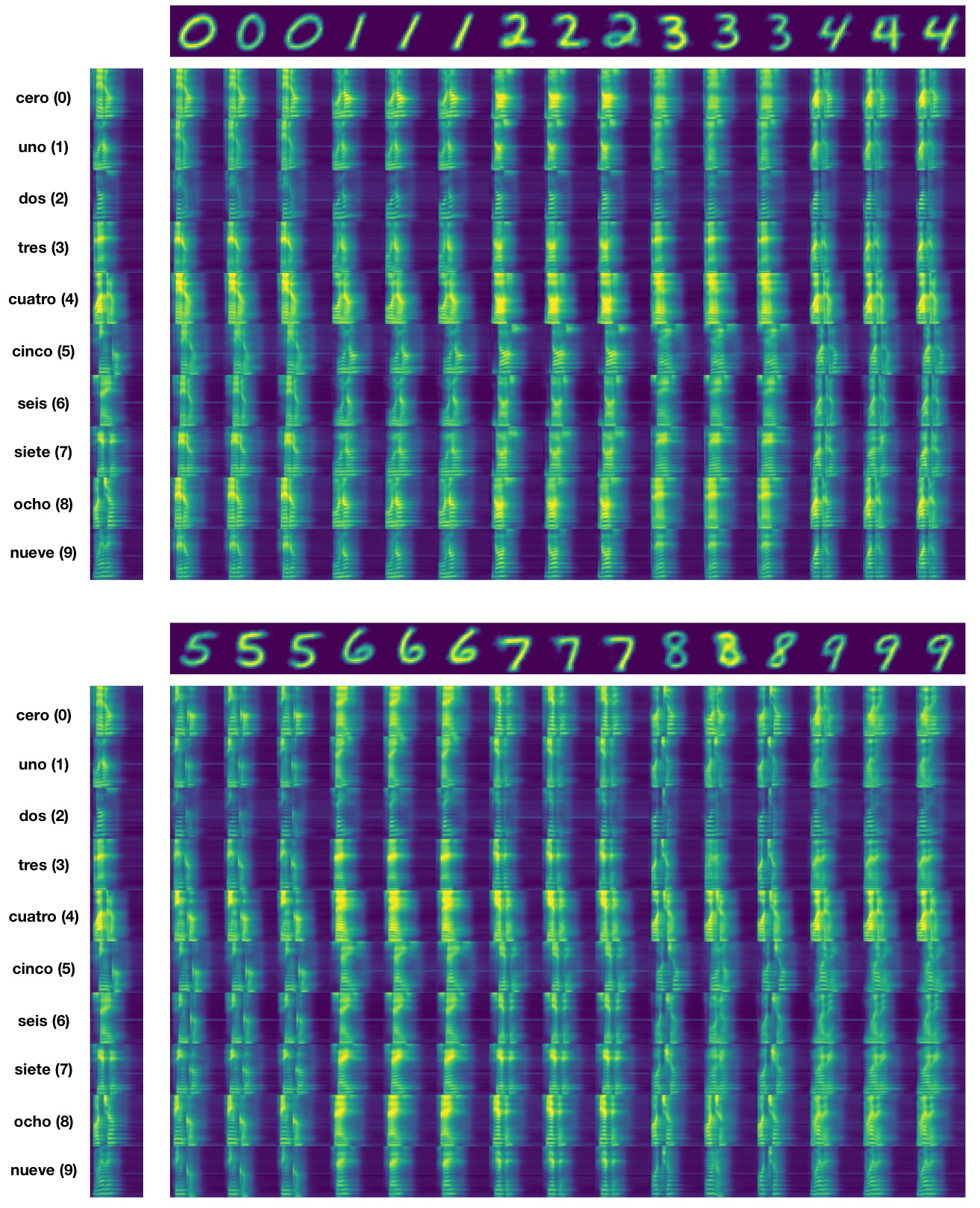}
  \caption{Generating spoken digits in Spanish from written digits. Samples within each column are conditioned on the same latent semantic variable $\z^s$ inferred from the image in the top row, and samples within each row are conditioned on the same latent audio style variable $\z^a$ inferred from the audio in the leftmost column. The digit identities are consistent within each column, and the speaking styles are consistent within each row.}
  \label{fig:spn_img2aud}
\end{figure}

\begin{figure}[ht]
  \centering
  \includegraphics[width=\linewidth]{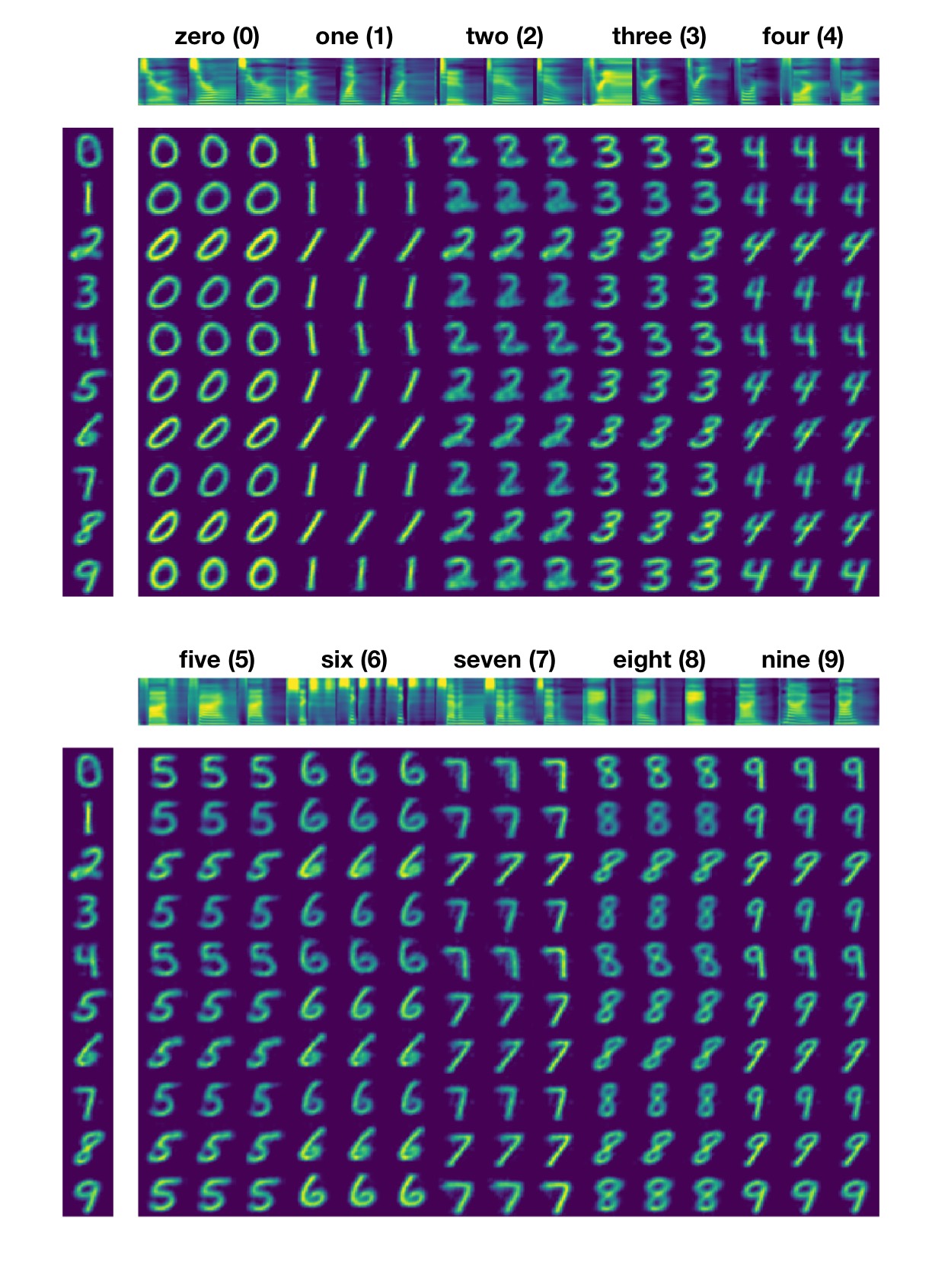}
  \caption{Generating written digits from spoken digits in English. Samples within each column are conditioned on the same latent semantic variable $\z^s$ inferred from the audio in the top row, and samples within each row are conditioned on the same latent image style variable $\z^i$ inferred from the image in the leftmost column. The digit identities are consistent within each column, and the writing styles are consistent within each row.}
  \label{fig:eng_aud2img}
\end{figure}

\begin{figure}[ht]
  \centering
  \includegraphics[width=\linewidth]{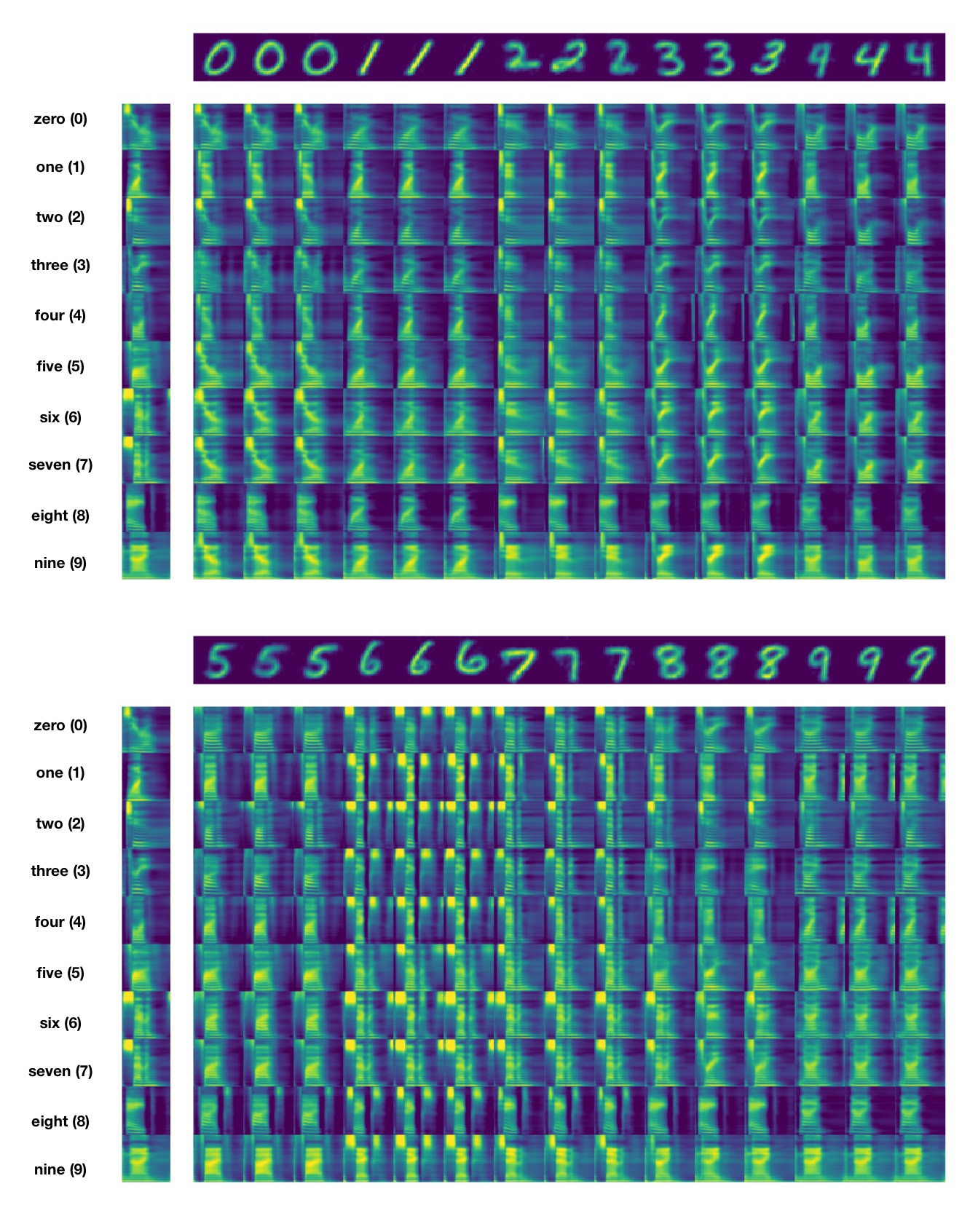}
  \caption{Generating spoken digits in English from written digits. Samples within each column are conditioned on the same latent semantic variable $\z^s$ inferred from the image in the top row, and samples within each row are conditioned on the same latent audio style variable $\z^a$ inferred from the audio in the leftmost column. The digit identities are consistent within each column, and the speaking styles are consistent within each row.}
  \label{fig:eng_img2aud}
\end{figure}

Results of conditional generation between Spanish spoken digits and written digits are shown in Figures~\ref{fig:spn_aud2img} and \ref{fig:spn_img2aud}, using the model trained on SecuVoice-MNIST.
Figures~\ref{fig:eng_aud2img} and \ref{fig:eng_img2aud} show the results of conditional generation between English spoken digits and written digits, using the model trained on TIDIGIT-MNIST.
All four figures demonstrate successful conditional generation and verify the ability of our model to learn disentangled representations for synthesizing novel samples.

\end{document}